\documentstyle[aclap]{article}

\author{David Elworthy\\
Sharp Laboratories of Europe Ltd. \\
Edmund Halley Road\\
Oxford Science Park\\
Oxford OX4 4GA\\
United Kingdom\\
{\tt dahe@sharp.co.uk}}
\title{Does Baum-Welch Re-estimation Help Taggers?}

\begin{document}
\maketitle
\begin{abstract}
In part of speech tagging by Hidden Markov Model, a statistical model is used
to assign grammatical categories to words in a text. Early work in the field
relied on a corpus which had been tagged by a human annotator to train the
model. More recently, Cutting {\it et al.} (1992) suggest that training can be
achieved with a minimal lexicon and a limited amount of {\em a priori}
information about probabilities, by using Baum-Welch re-estimation to
automatically refine the model. In this paper, I report two experiments
designed to determine how much manual training information is needed. The
first experiment suggests that initial biasing of either lexical or transition
probabilities is essential to achieve a good accuracy. The second experiment
reveals that there are three distinct patterns of Baum-Welch re-estimation. In
two of the patterns, the re-estimation ultimately reduces the accuracy of the
tagging rather than improving it. The pattern which is applicable can be
predicted from the quality of the initial model and the similarity between the
tagged training corpus (if any) and the corpus to be tagged. Heuristics for
deciding how to use re-estimation in an effective manner are given. The
conclusions are broadly in agreement with those of Merialdo (1994), but give
greater detail about the contributions of different parts of the model.

{\em Citation details: appears in Proceedings of the 4th ACL Conference on
Applied Natural Language Processing, Stuttgart, October 13-15th 1994, pp.
53--58. Some typos corrected from the published version.}
\end{abstract}
\bibliographystyle{acl}


\section{Background}

Part-of-speech tagging is the process of assigning grammatical categories to
individual words in a corpus. One widely used approach makes use of a
statistical technique called a Hidden Markov Model (HMM). The model is defined
by two collections of parameters: the {\em transition probabilities}, which
express the probability that a tag follows the preceding one (or two for a
second order model); and the {\em lexical probabilities}, giving the
probability that a word has a given tag without regard to words on either side
of it. To tag a text, the tags with non-zero probability are hypothesised for
each word, and the most probable sequence of tags given the sequence of words
is determined from the probabilities. Two algorithms are commonly used, known
as the Forward-Backward (FB) and Viterbi algorithms. FB assigns a probability
to every tag on every word, while Viterbi prunes tags which cannot be chosen
because their probability is lower than the ones of competing hypotheses, with
a corresponding gain in computational efficiency. For an introduction to the
algorithms, see Cutting {\it et al.} (1992), or the lucid description by
Sharman (1990).
 
There are two principal sources for the parameters of the model. If a tagged
corpus prepared by a human annotator is available, the transition and lexical
probabilities can be estimated from the frequencies of pairs of tags and of
tags associated with words. Alternatively, a procedure called Baum-Welch (BW)
re-estimation may be used, in which an untagged corpus is passed through the
FB algorithm with some initial model, and the resulting probabilities used to
determine new values for the lexical and transition probabilities. By
iterating the algorithm with the same corpus, the parameters of the model can
be made to converge on values which are locally optimal for the given
text. The degree of convergence can be measured using a perplexity measure,
the sum of \mbox{$p$log$_{2}p$} for hypothesis probabilities $p$, which gives
an estimate of the degree of disorder in the model.  The algorithm is again
described by Cutting {\it et al.} and by Sharman, and a mathematical
justification for it can be found in Huang {\it et al.} (1990).

The first major use of HMMs for part of speech tagging was in CLAWS
\cite{Garside:LAE} in the 1970s. With the availability of large corpora and
fast computers, there has been a recent resurgence of interest, and a number
of variations on and alternatives to the FB, Viterbi and BW algorithms have
been tried; see the work of, for example, Church \cite{Church:tag}, Brill
\cite{Brill:super,Brill:tag}, DeRose \cite{DeRose:disambig} and Kupiec
\cite{Kupiec:pos}. One of the most effective taggers based on a pure HMM is
that developed at Xerox \cite{Cutting:pos}. An important aspect of this tagger
is that it will give good accuracy with a minimal amount of manually tagged
training data. 96\% accuracy correct assignment of tags to word token,
compared with a human annotator, is quoted, over a 500000 word corpus.

The Xerox tagger attempts to avoid the need for a hand-tagged training corpus
as far as possible. Instead, an approximate model is constructed by hand,
which is then improved by BW re-estimation on an untagged training corpus. In
the above example, 8 iterations were sufficient. The initial model set up so
that some transitions and some tags in the lexicon are favoured, and hence
having a higher initial probability. Convergence of the model is improved by
keeping the number of parameters in the model down. To assist in this, low
frequency items in the lexicon are grouped together into equivalence classes,
such that all words in a given equivalence class have the same tags and
lexical probabilities, and whenever one of the words is looked up, then the
data common to all of them is used. Re-estimation on any of the words in a
class therefore counts towards re-estimation for all of them\footnote{The
technique was originally developed by Kupiec \cite{Kupiec:short}.}.

The results of the Xerox experiment appear very encouraging. Preparing tagged
corpora by hand is labour-intensive and potentially error-prone, and
although a semi-automatic approach can be used \cite{Marcus:penn}, it is a
good thing to reduce the human involvement as much as possible. However, some
careful examination of the experiment is needed. In the first place, Cutting
{\it et al.} do not compare the success rate in their work with that achieved
from a hand-tagged training text with no re-estimation. Secondly, it is
unclear how much the initial biasing contributes the success rate. If
significant human intervention is needed to provide the biasing, then the
advantages of automatic training become rather weaker, especially if such
intervention is needed on each new text domain. The kind of biasing Cutting
{\it et al.} describe reflects linguistic insights combined with an
understanding of the predictions a tagger could reasonably be expected to make
and the ones it could not.

The aim of this paper is to examine the role that training plays in the
tagging process, by an experimental evaluation of how the accuracy of the
tagger varies with the initial conditions. The results suggest that a
completely unconstrained initial model does not produce good quality results,
and that one accurately trained from a hand-tagged corpus will generally do
better than using an approach based on re-estimation, even when the training
comes from a different source. A second experiment shows that there are
different patterns of re-estimation, and that these patterns vary more or less
regularly with a broad characterisation of the initial conditions. The outcome
of the two experiments together points to heuristics for making effective use
of training and re-estimation, together with some directions for further
research.

Work similar to that described here has been carried out by Merialdo (1994),
with broadly similar conclusions. We will discuss this work below. The
principal contribution of this work is to separate the effect of the lexical
and transition parameters of the model, and to show how the results vary with
different degree of similarity between the training and test data.

\section{The tagger and corpora}

The experiments were conducted using two taggers, one written in C at
Cambridge University Computer Laboratory, and the other in C++ at Sharp
Laboratories.  Both taggers implement the FB, Viterbi and BW algorithms. For
training from a hand-tagged corpus, the model is estimated by counting the
number of transitions from each tag $i$ to each tag $j$, the total occurrence
of each tag $i$, and the total occurrence of word $w$ with tag $i$. Writing
these as $f(i,j)$, $f(i)$ and $f(i,w)$ respectively, the transition
probability from tag $i$ to tag $j$ is estimated as $f(i,j)/f(i)$ and the
lexical probability as $f(i,w)/f(i)$. Other estimation formulae have been used
in the past. For example, CLAWS \cite{Garside:LAE} normalises the lexical
probabilities by the total frequency of the word rather than of the
tag. Consulting the Baum-Welch re-estimation formulae suggests that the
approach described is more appropriate, and this is confirmed by slightly
greater tagging accuracy. Any transitions not seen in the training corpus are
given a small, non-zero probability.

The lexicon lists, for each word, all of tags seen in the training corpus with
their probabilities. For words not found in the lexicon, all open-class tags
are hypothesised, with equal probabilities. These words are added to the
lexicon at the end of first iteration when re-estimation is being used, so
that the probabilities of their hypotheses subsequently diverge from being
uniform.

To measure the accuracy of the tagger, we compare the chosen tag with one
provided by a human annotator. Various methods of quoting accuracy have been
used in the literature, the most common being the proportion of words (tokens)
receiving the correct tag. A better measure is the proportion of {\em
ambiguous} words which are given the correct tag, where by ambiguous we mean
that more than one tag was hypothesised. The former figure looks more
impressive, but the latter gives a better measure of how well the tagger is
doing, since it factors out the trivial assignment of tags to non-ambiguous
words. For a corpus in which a fraction $a$ of the words are ambiguous, and
$p$ is the accuracy on ambiguous words, the overall accuracy can be recovered
from \mbox{$1-a+pa$}. All of the accuracy figures quoted below are for
ambiguous words only.

The training and test corpora were drawn from the LOB corpus and the Penn
treebank. The hand tagging of these corpora is quite different. For example,
the LOB tagset used 134 tags, while the Penn treebank tagset has 48. The
general pattern of the results presented does not vary greatly with the corpus
and tagset used.

\section{The effect of the initial conditions}

The first experiment concerned the effect of the initial conditions on the
accuracy using Baum-Welch re-estimation. A model was trained from a
hand-tagged corpus in the manner described above, and then degraded in various
ways to simulate the effect of poorer training, as follows:
\begin{description}
\item[Lexicon]\mbox{}
\begin{description}
\item[D0] Un-degraded lexical probabilities, calculated from $f(i,w)/f(i)$.
\item[D1] Lexical probabilities are correctly ordered, so that the most
frequent tag has the highest lexical probability and so on, but the absolute
values are otherwise unreliable.
\item[D2] Lexical probabilities are proportional to the overall tag
frequencies, and are hence independent of the actual occurrence of the word in
the training corpus.
\item[D3] All lexical probabilities have the same value, so that the 
lexicon contains no information other than the possible tags for each word.
\end{description}
\item[Transitions]\mbox{}
\begin{description}
\item[T0] Un-degraded transition probabilities, calculated from $f(i,j)/f(i)$.
\item[T1] All transition probabilities have the same value.
\end{description}
\end{description}
We could expect to achieve D1 from, say, a printed dictionary listing parts of
speech in order of frequency. Perfect training is represented by case D0+T0.
The Xerox experiments \cite{Cutting:pos} correspond to something between D1
and D2, and between T0 and T1, in that there is some initial biasing of the
probabilities.

For the test, four corpora were constructed from the LOB corpus: LOB-B from
part B, LOB-L from part L, LOB-B-G from parts B to G inclusive and LOB-B-J
from parts B to J inclusive. Corpus LOB-B-J was used to train the model, and
LOB-B, LOB-L and LOB-B-G were passed through thirty iterations of the BW
algorithm as untagged data. In each case, the best accuracy (on ambiguous
words, as usual) from the FB algorithm was noted. As an additional test, we
tried assigning the most probable tag from the D0 lexicon, completely ignoring
tag-tag transitions. The results are summarised in table~\ref{tab-degrade},
for various corpora, where {\em F} denotes the ``most frequent tag'' test.
\begin{table*}[htb]
\begin{center}
\caption{Accuracy using Baum-Welch re-estimation with various initial
conditions}
\begin{tabular}{|c|c|c|c|c|} \hline
Dict & Trans & LOB-B (\%) & LOB-L (\%) & LOB-B-G (\%) \\ \hline
D0   & T0    & 95.96 & 94.77 & 96.17 \\
D1   & T0    & 95.40 & 94.44 & 95.40 \\
D2   & T0    & 90.52 & 91.82 & 92.36 \\
D3   & T0    & 92.96 & 92.80 & 93.48 \\
D0   & T1    & 94.06 & 92.27 & 94.51 \\
D1   & T1    & 94.06 & 92.27 & 94.51 \\
D2   & T1    & 66.51 & 72.48 & 55.88 \\
D3   & T1    & 75.49 & 80.87 & 79.12 \\
F    & -     & 89.22 & 85.32 & 88.71 \\ \hline
\end{tabular}
\label{tab-degrade}
\end{center}
\end{table*}
As an example of how these figures relate to overall accuracies, LOB-B
contains 32.35\% ambiguous tokens with respect to the lexicon from LOB-B-J,
and the overall accuracy in the D0+T0 case is hence 98.69\%. The general
pattern of the results is similar across the three test corpora, with the only
difference of interest being that case D3+T0 does better for LOB-L than for
the other two cases, and in particular does better than cases D0+T1 and D1+T1.
A possible explanation is that in this case the test data does not overlap
with the training data, and hence the good quality lexicons (D0 and D1) have
less of an influence. It is also interesting that D3+T1 does better than
D2+T1. The reasons for this are unclear, and the results are not always the
same with other corpora, which suggests that they are not statistically
significant.

Several follow-up experiments were used to confirm the results: using corpora
from the Penn treebank, using equivalence classes to ensure that all lexical
entries have a total relative frequency of at least 0.01, and using larger
corpora. The specific accuracies were different in the various tests, but the
overall patterns remained much the same, suggesting that they are not an
artifact of the tagset or of details of the text.

The observations we can make about these results are as follows. Firstly, two
of the tests, D2+T1 and D3+T1, give very poor performance. Their accuracy is
not even as good as that achieved by picking the most frequent tag (although
this of course implies a lexicon of D0 or D1 quality). It follows that if
Baum-Welch re-estimation is to be an effective technique, the initial data
must have either biasing in the transitions (the T0 cases) or in the lexical
probabilities (cases D0+T1 and D1+T1), but it is not necessary to have both
(D2/D3+T0 and D0/D1+T1).

Secondly, training from a hand-tagged corpus (case D0+T0) always does best,
even when the test data is from a different source to the training data, as it
is for LOB-L. So perhaps it is worth investing effort in hand-tagging training
corpora after all, rather than just building a lexicon and letting
re-estimation sort out the probabilities. But how can we ensure that
re-estimation will produce a good quality model? We look further at this issue
in the next section.

\section{Patterns of re-estimation}

\begin{figure*}
\begin{center}
\item[]
\setlength{\unitlength}{0.240900pt}
\ifx\plotpoint\undefined\newsavebox{\plotpoint}\fi
\sbox{\plotpoint}{\rule[-0.200pt]{0.400pt}{0.400pt}}%
\begin{picture}(1500,900)(0,0)
\font\gnuplot=cmr10 at 10pt
\gnuplot
\sbox{\plotpoint}{\rule[-0.200pt]{0.400pt}{0.400pt}}%
\put(220.0,113.0){\rule[-0.200pt]{0.400pt}{184.048pt}}
\put(220.0,113.0){\rule[-0.200pt]{4.818pt}{0.400pt}}
\put(198,113){\makebox(0,0)[r]{78}}
\put(1416.0,113.0){\rule[-0.200pt]{4.818pt}{0.400pt}}
\put(220.0,209.0){\rule[-0.200pt]{4.818pt}{0.400pt}}
\put(198,209){\makebox(0,0)[r]{80}}
\put(1416.0,209.0){\rule[-0.200pt]{4.818pt}{0.400pt}}
\put(220.0,304.0){\rule[-0.200pt]{4.818pt}{0.400pt}}
\put(198,304){\makebox(0,0)[r]{82}}
\put(1416.0,304.0){\rule[-0.200pt]{4.818pt}{0.400pt}}
\put(220.0,400.0){\rule[-0.200pt]{4.818pt}{0.400pt}}
\put(198,400){\makebox(0,0)[r]{84}}
\put(1416.0,400.0){\rule[-0.200pt]{4.818pt}{0.400pt}}
\put(220.0,495.0){\rule[-0.200pt]{4.818pt}{0.400pt}}
\put(198,495){\makebox(0,0)[r]{86}}
\put(1416.0,495.0){\rule[-0.200pt]{4.818pt}{0.400pt}}
\put(220.0,591.0){\rule[-0.200pt]{4.818pt}{0.400pt}}
\put(198,591){\makebox(0,0)[r]{88}}
\put(1416.0,591.0){\rule[-0.200pt]{4.818pt}{0.400pt}}
\put(220.0,686.0){\rule[-0.200pt]{4.818pt}{0.400pt}}
\put(198,686){\makebox(0,0)[r]{90}}
\put(1416.0,686.0){\rule[-0.200pt]{4.818pt}{0.400pt}}
\put(220.0,782.0){\rule[-0.200pt]{4.818pt}{0.400pt}}
\put(198,782){\makebox(0,0)[r]{92}}
\put(1416.0,782.0){\rule[-0.200pt]{4.818pt}{0.400pt}}
\put(220.0,877.0){\rule[-0.200pt]{4.818pt}{0.400pt}}
\put(198,877){\makebox(0,0)[r]{94}}
\put(1416.0,877.0){\rule[-0.200pt]{4.818pt}{0.400pt}}
\put(220.0,113.0){\rule[-0.200pt]{0.400pt}{4.818pt}}
\put(220,68){\makebox(0,0){0}}
\put(220.0,857.0){\rule[-0.200pt]{0.400pt}{4.818pt}}
\put(423.0,113.0){\rule[-0.200pt]{0.400pt}{4.818pt}}
\put(423,68){\makebox(0,0){5}}
\put(423.0,857.0){\rule[-0.200pt]{0.400pt}{4.818pt}}
\put(625.0,113.0){\rule[-0.200pt]{0.400pt}{4.818pt}}
\put(625,68){\makebox(0,0){10}}
\put(625.0,857.0){\rule[-0.200pt]{0.400pt}{4.818pt}}
\put(828.0,113.0){\rule[-0.200pt]{0.400pt}{4.818pt}}
\put(828,68){\makebox(0,0){15}}
\put(828.0,857.0){\rule[-0.200pt]{0.400pt}{4.818pt}}
\put(1031.0,113.0){\rule[-0.200pt]{0.400pt}{4.818pt}}
\put(1031,68){\makebox(0,0){20}}
\put(1031.0,857.0){\rule[-0.200pt]{0.400pt}{4.818pt}}
\put(1233.0,113.0){\rule[-0.200pt]{0.400pt}{4.818pt}}
\put(1233,68){\makebox(0,0){25}}
\put(1233.0,857.0){\rule[-0.200pt]{0.400pt}{4.818pt}}
\put(1436.0,113.0){\rule[-0.200pt]{0.400pt}{4.818pt}}
\put(1436,68){\makebox(0,0){30}}
\put(1436.0,857.0){\rule[-0.200pt]{0.400pt}{4.818pt}}
\put(220.0,113.0){\rule[-0.200pt]{292.934pt}{0.400pt}}
\put(1436.0,113.0){\rule[-0.200pt]{0.400pt}{184.048pt}}
\put(220.0,877.0){\rule[-0.200pt]{292.934pt}{0.400pt}}
\put(45,495){\makebox(0,0){Accuracy}}
\put(65,455){\makebox(0,0){(\%)}}
\put(828,23){\makebox(0,0){Iteration}}
\put(220.0,113.0){\rule[-0.200pt]{0.400pt}{184.048pt}}
\put(1306,812){\makebox(0,0)[r]{Initial}}
\put(1328.0,812.0){\rule[-0.200pt]{15.899pt}{0.400pt}}
\put(261,813){\usebox{\plotpoint}}
\multiput(261.58,809.31)(0.498,-0.991){77}{\rule{0.120pt}{0.890pt}}
\multiput(260.17,811.15)(40.000,-77.153){2}{\rule{0.400pt}{0.445pt}}
\multiput(301.00,732.92)(0.499,-0.498){79}{\rule{0.500pt}{0.120pt}}
\multiput(301.00,733.17)(39.962,-41.000){2}{\rule{0.250pt}{0.400pt}}
\multiput(342.00,691.92)(0.646,-0.497){59}{\rule{0.616pt}{0.120pt}}
\multiput(342.00,692.17)(38.721,-31.000){2}{\rule{0.308pt}{0.400pt}}
\multiput(382.00,660.92)(0.896,-0.496){43}{\rule{0.813pt}{0.120pt}}
\multiput(382.00,661.17)(39.312,-23.000){2}{\rule{0.407pt}{0.400pt}}
\multiput(423.00,637.92)(1.122,-0.495){33}{\rule{0.989pt}{0.119pt}}
\multiput(423.00,638.17)(37.948,-18.000){2}{\rule{0.494pt}{0.400pt}}
\multiput(463.00,619.92)(1.386,-0.494){27}{\rule{1.193pt}{0.119pt}}
\multiput(463.00,620.17)(38.523,-15.000){2}{\rule{0.597pt}{0.400pt}}
\multiput(504.00,604.92)(1.352,-0.494){27}{\rule{1.167pt}{0.119pt}}
\multiput(504.00,605.17)(37.579,-15.000){2}{\rule{0.583pt}{0.400pt}}
\multiput(544.00,589.92)(1.607,-0.493){23}{\rule{1.362pt}{0.119pt}}
\multiput(544.00,590.17)(38.174,-13.000){2}{\rule{0.681pt}{0.400pt}}
\multiput(585.00,576.94)(5.745,-0.468){5}{\rule{4.100pt}{0.113pt}}
\multiput(585.00,577.17)(31.490,-4.000){2}{\rule{2.050pt}{0.400pt}}
\multiput(625.00,572.93)(3.655,-0.482){9}{\rule{2.833pt}{0.116pt}}
\multiput(625.00,573.17)(35.119,-6.000){2}{\rule{1.417pt}{0.400pt}}
\multiput(666.00,566.93)(4.384,-0.477){7}{\rule{3.300pt}{0.115pt}}
\multiput(666.00,567.17)(33.151,-5.000){2}{\rule{1.650pt}{0.400pt}}
\put(706,561.17){\rule{8.300pt}{0.400pt}}
\multiput(706.00,562.17)(23.773,-2.000){2}{\rule{4.150pt}{0.400pt}}
\multiput(747.00,559.93)(2.607,-0.488){13}{\rule{2.100pt}{0.117pt}}
\multiput(747.00,560.17)(35.641,-8.000){2}{\rule{1.050pt}{0.400pt}}
\multiput(787.00,551.92)(1.488,-0.494){25}{\rule{1.271pt}{0.119pt}}
\multiput(787.00,552.17)(38.361,-14.000){2}{\rule{0.636pt}{0.400pt}}
\multiput(828.00,537.93)(2.673,-0.488){13}{\rule{2.150pt}{0.117pt}}
\multiput(828.00,538.17)(36.538,-8.000){2}{\rule{1.075pt}{0.400pt}}
\multiput(869.00,529.95)(8.723,-0.447){3}{\rule{5.433pt}{0.108pt}}
\multiput(869.00,530.17)(28.723,-3.000){2}{\rule{2.717pt}{0.400pt}}
\put(909,526.17){\rule{8.300pt}{0.400pt}}
\multiput(909.00,527.17)(23.773,-2.000){2}{\rule{4.150pt}{0.400pt}}
\put(950,524.17){\rule{8.100pt}{0.400pt}}
\multiput(950.00,525.17)(23.188,-2.000){2}{\rule{4.050pt}{0.400pt}}
\multiput(990.00,522.95)(8.946,-0.447){3}{\rule{5.567pt}{0.108pt}}
\multiput(990.00,523.17)(29.446,-3.000){2}{\rule{2.783pt}{0.400pt}}
\put(1031,519.17){\rule{8.100pt}{0.400pt}}
\multiput(1031.00,520.17)(23.188,-2.000){2}{\rule{4.050pt}{0.400pt}}
\put(1071,517.67){\rule{9.877pt}{0.400pt}}
\multiput(1071.00,518.17)(20.500,-1.000){2}{\rule{4.938pt}{0.400pt}}
\put(1152,516.67){\rule{9.877pt}{0.400pt}}
\multiput(1152.00,517.17)(20.500,-1.000){2}{\rule{4.938pt}{0.400pt}}
\multiput(1193.00,515.94)(5.745,-0.468){5}{\rule{4.100pt}{0.113pt}}
\multiput(1193.00,516.17)(31.490,-4.000){2}{\rule{2.050pt}{0.400pt}}
\put(1233,511.67){\rule{9.877pt}{0.400pt}}
\multiput(1233.00,512.17)(20.500,-1.000){2}{\rule{4.938pt}{0.400pt}}
\multiput(1274.00,510.94)(5.745,-0.468){5}{\rule{4.100pt}{0.113pt}}
\multiput(1274.00,511.17)(31.490,-4.000){2}{\rule{2.050pt}{0.400pt}}
\put(1314,506.67){\rule{9.877pt}{0.400pt}}
\multiput(1314.00,507.17)(20.500,-1.000){2}{\rule{4.938pt}{0.400pt}}
\put(1355,505.17){\rule{8.100pt}{0.400pt}}
\multiput(1355.00,506.17)(23.188,-2.000){2}{\rule{4.050pt}{0.400pt}}
\multiput(1395.00,503.94)(5.891,-0.468){5}{\rule{4.200pt}{0.113pt}}
\multiput(1395.00,504.17)(32.283,-4.000){2}{\rule{2.100pt}{0.400pt}}
\put(1112.0,518.0){\rule[-0.200pt]{9.636pt}{0.400pt}}
\put(1306,767){\makebox(0,0)[r]{Early}}
\multiput(1328,767)(20.756,0.000){4}{\usebox{\plotpoint}}
\put(1394,767){\usebox{\plotpoint}}
\put(261,469){\usebox{\plotpoint}}
\multiput(261,469)(3.204,20.507){13}{\usebox{\plotpoint}}
\multiput(301,725)(13.810,-15.494){3}{\usebox{\plotpoint}}
\multiput(342,679)(13.619,-15.662){3}{\usebox{\plotpoint}}
\multiput(382,633)(15.409,-13.905){3}{\usebox{\plotpoint}}
\multiput(423,596)(8.670,-18.858){4}{\usebox{\plotpoint}}
\multiput(463,509)(17.140,-11.705){3}{\usebox{\plotpoint}}
\multiput(504,481)(19.102,-8.118){2}{\usebox{\plotpoint}}
\multiput(544,464)(20.273,-4.450){2}{\usebox{\plotpoint}}
\multiput(585,455)(19.590,-6.857){2}{\usebox{\plotpoint}}
\multiput(625,441)(20.047,-5.378){2}{\usebox{\plotpoint}}
\multiput(666,430)(20.595,-2.574){2}{\usebox{\plotpoint}}
\multiput(706,425)(18.832,-8.727){2}{\usebox{\plotpoint}}
\multiput(747,406)(20.352,-4.070){2}{\usebox{\plotpoint}}
\multiput(787,398)(20.700,-1.515){2}{\usebox{\plotpoint}}
\multiput(828,395)(20.537,-3.005){2}{\usebox{\plotpoint}}
\multiput(869,389)(20.526,-3.079){2}{\usebox{\plotpoint}}
\multiput(909,383)(20.459,-3.493){2}{\usebox{\plotpoint}}
\multiput(950,376)(19.880,-5.964){2}{\usebox{\plotpoint}}
\multiput(990,364)(20.700,-1.515){2}{\usebox{\plotpoint}}
\multiput(1031,361)(20.526,-3.079){2}{\usebox{\plotpoint}}
\multiput(1071,355)(20.700,-1.515){2}{\usebox{\plotpoint}}
\multiput(1112,352)(20.730,-1.036){2}{\usebox{\plotpoint}}
\multiput(1152,350)(20.731,-1.011){2}{\usebox{\plotpoint}}
\multiput(1193,348)(20.526,-3.079){2}{\usebox{\plotpoint}}
\multiput(1233,342)(20.749,-0.506){2}{\usebox{\plotpoint}}
\put(1293.87,337.52){\usebox{\plotpoint}}
\multiput(1314,334)(20.657,-2.015){2}{\usebox{\plotpoint}}
\multiput(1355,330)(20.730,-1.036){2}{\usebox{\plotpoint}}
\multiput(1395,328)(20.657,-2.015){2}{\usebox{\plotpoint}}
\put(1436,324){\usebox{\plotpoint}}
\sbox{\plotpoint}{\rule[-0.400pt]{0.800pt}{0.800pt}}%
\put(1306,722){\makebox(0,0)[r]{Classical}}
\put(1328.0,722.0){\rule[-0.400pt]{15.899pt}{0.800pt}}
\put(261,134){\usebox{\plotpoint}}
\multiput(262.41,134.00)(0.502,3.249){73}{\rule{0.121pt}{5.320pt}}
\multiput(259.34,134.00)(40.000,244.958){2}{\rule{0.800pt}{2.660pt}}
\multiput(302.41,390.00)(0.502,0.709){75}{\rule{0.121pt}{1.332pt}}
\multiput(299.34,390.00)(41.000,55.236){2}{\rule{0.800pt}{0.666pt}}
\multiput(342.00,449.41)(0.668,0.503){53}{\rule{1.267pt}{0.121pt}}
\multiput(342.00,446.34)(37.371,30.000){2}{\rule{0.633pt}{0.800pt}}
\multiput(382.00,479.40)(2.504,0.516){11}{\rule{3.844pt}{0.124pt}}
\multiput(382.00,476.34)(33.021,9.000){2}{\rule{1.922pt}{0.800pt}}
\put(423,487.34){\rule{8.200pt}{0.800pt}}
\multiput(423.00,485.34)(22.980,4.000){2}{\rule{4.100pt}{0.800pt}}
\put(463,488.34){\rule{9.877pt}{0.800pt}}
\multiput(463.00,489.34)(20.500,-2.000){2}{\rule{4.938pt}{0.800pt}}
\put(504,488.84){\rule{9.636pt}{0.800pt}}
\multiput(504.00,487.34)(20.000,3.000){2}{\rule{4.818pt}{0.800pt}}
\put(544,491.34){\rule{9.877pt}{0.800pt}}
\multiput(544.00,490.34)(20.500,2.000){2}{\rule{4.938pt}{0.800pt}}
\put(585,491.34){\rule{9.636pt}{0.800pt}}
\multiput(585.00,492.34)(20.000,-2.000){2}{\rule{4.818pt}{0.800pt}}
\put(625,490.84){\rule{9.877pt}{0.800pt}}
\multiput(625.00,490.34)(20.500,1.000){2}{\rule{4.938pt}{0.800pt}}
\put(666,490.34){\rule{9.636pt}{0.800pt}}
\multiput(666.00,491.34)(20.000,-2.000){2}{\rule{4.818pt}{0.800pt}}
\put(706,488.34){\rule{9.877pt}{0.800pt}}
\multiput(706.00,489.34)(20.500,-2.000){2}{\rule{4.938pt}{0.800pt}}
\put(747,486.84){\rule{9.636pt}{0.800pt}}
\multiput(747.00,487.34)(20.000,-1.000){2}{\rule{4.818pt}{0.800pt}}
\put(828,485.84){\rule{9.877pt}{0.800pt}}
\multiput(828.00,486.34)(20.500,-1.000){2}{\rule{4.938pt}{0.800pt}}
\put(787.0,488.0){\rule[-0.400pt]{9.877pt}{0.800pt}}
\put(950,484.34){\rule{9.636pt}{0.800pt}}
\multiput(950.00,485.34)(20.000,-2.000){2}{\rule{4.818pt}{0.800pt}}
\put(990,482.84){\rule{9.877pt}{0.800pt}}
\multiput(990.00,483.34)(20.500,-1.000){2}{\rule{4.938pt}{0.800pt}}
\put(1031,481.84){\rule{9.636pt}{0.800pt}}
\multiput(1031.00,482.34)(20.000,-1.000){2}{\rule{4.818pt}{0.800pt}}
\put(869.0,487.0){\rule[-0.400pt]{19.513pt}{0.800pt}}
\put(1112,480.84){\rule{9.636pt}{0.800pt}}
\multiput(1112.00,481.34)(20.000,-1.000){2}{\rule{4.818pt}{0.800pt}}
\multiput(1152.00,480.08)(2.504,-0.516){11}{\rule{3.844pt}{0.124pt}}
\multiput(1152.00,480.34)(33.021,-9.000){2}{\rule{1.922pt}{0.800pt}}
\put(1193,470.84){\rule{9.636pt}{0.800pt}}
\multiput(1193.00,471.34)(20.000,-1.000){2}{\rule{4.818pt}{0.800pt}}
\put(1233,469.84){\rule{9.877pt}{0.800pt}}
\multiput(1233.00,470.34)(20.500,-1.000){2}{\rule{4.938pt}{0.800pt}}
\put(1274,468.84){\rule{9.636pt}{0.800pt}}
\multiput(1274.00,469.34)(20.000,-1.000){2}{\rule{4.818pt}{0.800pt}}
\put(1071.0,483.0){\rule[-0.400pt]{9.877pt}{0.800pt}}
\put(1355,466.84){\rule{9.636pt}{0.800pt}}
\multiput(1355.00,468.34)(20.000,-3.000){2}{\rule{4.818pt}{0.800pt}}
\put(1314.0,470.0){\rule[-0.400pt]{9.877pt}{0.800pt}}
\put(1395.0,467.0){\rule[-0.400pt]{9.877pt}{0.800pt}}
\end{picture}
\caption{Example Baum-Welch behaviour}
\label{fig-beh}
\end{center}
\end{figure*}
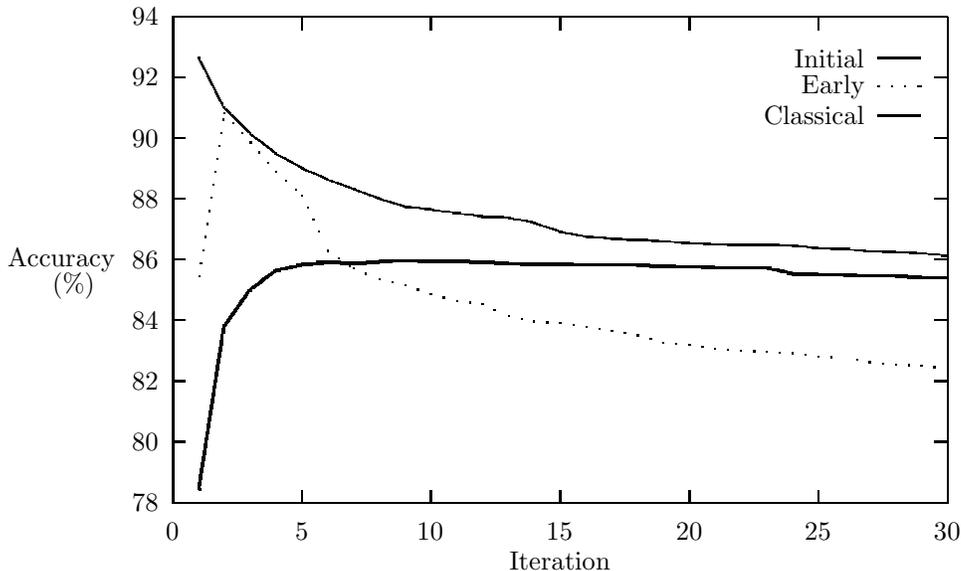

During the first experiment, it became apparent that Baum-Welch re-estimation
sometimes decreases the accuracy as the iteration progresses. A second
experiment was conducted to decide when it is appropriate to use Baum-Welch
re-estimation at all. There seem to be three patterns of behaviour:
\begin{description}
\item[Classical] A general trend of rising accuracy on each iteration, with
any falls in accuracy being local. It indicates that the model is converging
towards an optimum which is better than its starting point.
\item[Initial maximum] Highest accuracy on the first iteration, and falling
thereafter. In this case the initial model is of better quality than BW can
achieve. That is, while BW will converge on an optimum, the notion of
optimality is with respect to the HMM rather than to the linguistic judgements
about correct tagging.
\item[Early maximum] Rising accuracy for a small number of iterations (2--4),
and then falling as in initial maximum.
\end{description}
An example of each of the three behaviours is shown in figure~\ref{fig-beh}.
The values of the accuracies and the test conditions are unimportant here; all
we want to show is the general patterns.
The second experiment had the aim of trying to discover which pattern applies
under which circumstances, in order to help decide how to train the model.
Clearly, if the expected pattern is initial maximum, we should not use BW at
all, if early maximum, we should halt the process after a few iterations, and
if classical, we should halt the process in a ``standard'' way, such as
comparing the perplexity of successive models.

The tests were conducted in a similar manner to those of the first experiment,
by building a lexicon and transitions from a hand tagged training corpus, and
then applying them to a test corpus with varying degrees of degradation.
Firstly, four different degrees of degradation were used: no degradation at
all, D2 degradation of the lexicon, T1 degradation of the transitions, and the
two together. Secondly, we selected test corpora with varying degrees of
similarity to the training corpus: the same text, text from a similar domain,
and text which is significantly different. Two tests were conducted with each
combination of the degradation and similarity, using different corpora (from
the Penn treebank) ranging in size from approximately 50000 words to 500000
words. The re-estimation was allowed to run for ten iterations.

The results appear in table~\ref{tab-pat1}, showing the best accuracy achieved
(on ambiguous words).  the iteration at which it occurred, and the pattern
of re-estimation (\mbox{I = initial maximum}, \mbox{E = early maximum},
\mbox{C = classical}). The patterns are summarised in table~\ref{tab-pat2},
each entry in the table showing the patterns for the two tests under the given
conditions.
\begin{table*}[htb]
\begin{center}
\caption{Baum-Welch patterns (data)}
\begin{tabular}{|c|c|c|c|c|c|c|c|} \hline
Corpus   & Degradation & \multicolumn{3}{c|}{Test 1} &
\multicolumn{3}{c|}{Test 2} \\
\cline{3-5} \cline{6-8}
relation &             & Best (\%) & at & pattern & Best (\%) & at & pattern \\ \hline
Same      & D0+T0 & 93.11 & 1 & I & 92.83 & 1 & I \\
Similar   & D0+T0 & 89.95 & 1 & I & 75.03 & 2 & E \\
Different & D0+T0 & 84.59 & 2 & E & 86.00 & 2 & E \\ \hline
Same      & D0+T1 & 91.71 & 2 & E & 90.52 & 2 & E \\
Similar   & D0+T1 & 87.93 & 2 & E & 70.63 & 3 & E \\
Different & D0+T1 & 80.87 & 3 & E & 82.68 & 3 & E \\ \hline
Same      & D2+T0 & 84.87 & 10 & C & 87.31 & 8 & C \\
Similar   & D2+T0 & 81.07 & 9 & C & 71.40 & 4 & C$^{*}$ \\
Different & D2+T0 & 78.54 & 5 & C$^{*}$ & 80.81 & 9 & C \\ \hline
Same      & D2+T1 & 72.58 & 9 & C & 80.53 & 10 & C \\
Similar   & D2+T1 & 68.35 & 10 & C & 62.76 & 10 & C \\
Different & D2+T1 & 65.64 & 10 & C & 68.95 & 10 & C \\ \hline
\end{tabular}
\newline
$^{*}$ These tests gave an early peak, but the graphs of accuracy against
number of iterations show the pattern to be classical rather than early
maximum.
\label{tab-pat1}
\end{center}
\end{table*}
\begin{table*}[htb]
\begin{center}
\caption{Baum-Welch patterns (summary)}
\begin{tabular}{|c|c|c|c|c|} \hline
{\em Degradation}     & D0+T0 & D0+T1 & D2+T0 & D2+T1 \\ \hline
{\em Corpus relation} &       &       &       &       \\
Same            & I, I  & E, E  & C, C  & C, C \\
Similar         & I, E  & E, E  & C, C  & C, C \\
Different       & E, E  & E, E  & C, C  & C, C \\ \hline
\end{tabular}
\label{tab-pat2}
\end{center}
\end{table*}
Although there is some variations in the readings, for example in the
``similar/D0+T0'' case, we can draw some general conclusions about the
patterns obtained from different sorts of data. When the lexicon is degraded
(D2), the pattern is always classical. With a good lexicon but either degraded
transitions or a test corpus differing from the training corpus, the pattern
tends to be early maximum. When the test corpus is very similar to the model,
then the pattern is initial maximum. Furthermore, examining the accuracies in
table~\ref{tab-pat1}, in the cases of initial maximum and early maximum, the
accuracy tends to be significantly higher than with classical behaviour. It
seems likely that what is going on is that the model is converging to towards
something of similar ``quality'' in each case, but when the pattern is
classical, the convergence starts from a lower quality model and improves, and
in the other cases, it starts from a higher quality one and deteriorates.  In
the case of early maximum, the few iterations where the accuracy is improving
correspond to the creation of entries for unknown words and the fine tuning of
ones for known ones, and these changes outweigh those produced by the
re-estimation.

\section{Discussion}

From the observations in the previous section, we propose the following
guidelines for how to train a HMM for use in tagging:
\begin{itemize}
\item If a hand-tagged training corpus is available, use it
.
If the test and training
corpora are near-identical, do not use BW re-estimation;
otherwise use for a small number of iterations.
\item If no such training corpus is available, but a lexicon with at least
relative frequency data is available, use BW re-estimation for a small number
of iterations.
\item If neither training corpus nor lexicon are available, use BW
re-estimation with standard convergence tests such as perplexity.
Without a lexicon, some initial biasing of the transitions is needed if good
results are to be obtained.
\end{itemize}

Similar results are presented by Merialdo (1994), who describes experiments to
compare the effect of training from a hand-tagged corpora and using the
Baum-Welch algorithm with various initial conditions.  As in the experiments
above, BW re-estimation gave a decrease in accuracy when the starting point
was derived from a significant amount of hand-tagged text. In addition,
although Merialdo does not highlight the point, BW re-estimation starting from
less than 5000 words of hand-tagged text shows early maximum behaviour.
Merialdo's conclusion is that taggers should be trained using as much
hand-tagged text as possible to begin with, and only then applying BW
re-estimation with untagged text. The step forward taken in the work here is
to show that there are three patterns of re-estimation behaviour, with
differing guidelines for how to use BW effectively, and that to obtain a good
starting point when a hand-tagged corpus is not available or is too small,
either the lexicon or the transitions must be biased.

While these may be useful heuristics from a practical point of view, the next
step forward is to look for an automatic way of predicting the accuracy of the
tagging process given a corpus and a model.  Some preliminary experiments with
using measures such as perplexity and the average probability of hypotheses
show that, while they do give an indication of convergence during
re-estimation, neither shows a strong correlation with the accuracy. Perhaps
what is needed is a ``similarity measure'' between two models $M$ and $M'$,
such that if a corpus were tagged with model $M$, $M'$ is the model obtained
by training from the output corpus from the tagger as if it were a hand-tagged
corpus. However, preliminary experiments using such measures as the
Kullback-Liebler distance between the initial and new models have again showed
that it does not give good predictions of accuracy. In the end it may turn out
there is simply no way of making the prediction without a source of
information extrinsic to both model and corpus.

\section*{Acknowledgements}

The work described here was carried out at the Cambridge University Computer
Laboratory as part of Esprit BR Project 7315 ``The Acquisition of Lexical
Knowledge'' (Acquilex-II). The results were confirmed and extended at Sharp
Laboratories of Europe. I thank Ted Briscoe for his guidance and advice, 
and the ANLP referees for their comments.


\end{document}